\documentclass{fairmeta}

\definecolor{mycolor}{HTML}{7A4AC8}

\usepackage{template/macro}
\input{template/package}
\usepackage{xspace}
\usepackage{todonotes}

\providecommand{\Mathlib}{\textsc{mathlib}\xspace}
\providecommand{\Atlas}{\textsc{Atlas}\xspace}  

\newcommand{\lgp}[1]{\ensuremath{\log(1+#1)}}      

\newcommand{\Deps}{\ensuremath{\operatorname{deps}}}
\newcommand{\Nlines}{\ensuremath{\operatorname{lines}}}
\newcommand{\role}[1]{\text{\texttt{#1}}}   

\makeatletter
\@ifpackageloaded{algpseudocode}{}{%
  \IfFileExists{algorithm.sty}{%
    \usepackage{algorithm}%
    \usepackage{algorithmicx}%
    \usepackage{algpseudocode}%
  }{}%
}
\makeatother

\makeatletter
\@ifpackageloaded{algpseudocode}{%
  \algrenewcommand{\algorithmiccomment}[1]{\hfill\textcolor{mwGray}{\footnotesize$\triangleright$~#1}}%
  \algrenewcommand\algorithmicrequire{\textbf{Input:}}%
  \algrenewcommand\algorithmicensure{\textbf{Output:}}%
  \algnewcommand{\LineComment}[1]{\State\textcolor{mwGray}{\footnotesize$\triangleright$~#1}}%
  \algnewcommand{\Fn}[1]{\textsc{#1}}%
}{}
\makeatother

\newif\ifphshow
\phshowtrue
\definecolor{mwPlaceholder}{HTML}{9B5DE5}

\usetikzlibrary{arrows.meta,positioning,shapes.geometric,calc,fit,backgrounds,matrix}

\definecolor{mwInk}{HTML}{252525}
\definecolor{mwBlue}{HTML}{2C6E9F}
\definecolor{mwTeal}{HTML}{2A8178}
\definecolor{mwGold}{HTML}{B48622}
\definecolor{mwRed}{HTML}{C44E3B}
\definecolor{mwViolet}{HTML}{7666A5}
\definecolor{mwGray}{HTML}{6B6B6B}
\pgfplotsset{
  mwplot/.style={
    axis line style={mwInk!65},
    tick style={mwInk!55},
    tick label style={font=\scriptsize,text=mwInk},
    label style={font=\footnotesize,text=mwInk},
    grid style={mwInk!8},
    legend style={font=\scriptsize,draw=mwInk!18,fill=white,rounded corners=1pt},
  }
}
\tikzset{
  mwbox/.style={draw=mwInk!35,fill=white,rounded corners=2pt,thick,
    align=center,inner xsep=7pt,inner ysep=5pt,font=\small,text=mwInk},
  mwdata/.style={mwbox,draw=mwBlue!60,fill=mwBlue!7},
  mwmodel/.style={mwbox,draw=mwGold!70,fill=mwGold!9},
  mwresult/.style={mwbox,draw=mwTeal!70,fill=mwTeal!8},
  mwtarget/.style={mwbox,draw=mwViolet!65,fill=mwViolet!8},
  mwflow/.style={-{Stealth[length=2.1mm]},thick,draw=mwInk!60},
  mwlabel/.style={font=\scriptsize,text=mwGray,align=center},
}

\providecommand{\note}[1]{}
\renewcommand{\note}[1]{}

\title{Learning to Discover Interesting Mathematics}

\author[1, 2 *]{Niket Patel}
\author[1, 3]{Ahmad Rammal}
\author[3]{Amaury Hayat}
\author[1]{Remi Munos}
\author[1, 2]{Julia Kempe}

\affiliation[1]{FAIR @ Meta}
\affiliation[2]{New York University}
\affiliation[3]{CERMICS, ENPC, Institut Polytechnique de Paris}

\contribution[*]{Work done while at Meta.}

\abstract{Recently, Large Language Models (LLMs) have been increasingly able to solve advanced mathematical problems, including many that have been open for decades.
This opens the door to expansion of mathematical knowledge at unprecedented scale.
Yet, while LLMs may be able to conjecture and prove more and more theorems, it remains open whether this new mathematical knowledge is \emph{interesting} or \emph{useful}.
We define intrinsic interestingness of a theorem as the ratio between the length of its proof and the length of its statement. We show that this correlates strongly with an extrinsic measure of the downstream utility of a theorem.
We identify the difficulty of a proof conditioned on a set of premises as a useful primitive for computing these metrics, and train a 27B model that predicts proof difficulty more accurately than frontier general-purpose models.
Optimizing for our metric creates a model capable of producing more interesting theorems, while also reducing substantial or full overlap with \Mathlib from $91.9\%$ to $30.6\%$, showcasing the creation of more out-of-distribution math.
We show that our system can generate candidate theorems, select the most interesting among them, and iteratively build on a self-expanding mathematical library.
These metrics provide a practical and quantifiable signal for ranking conjectures and guiding proof search within formal mathematical libraries.
Our framework provides a path towards self-expanding, machine-verified mathematical libraries that can choose worthwhile statements without relying on human-supplied targets.
}

\date{\today}
\correspondence{Correspondence to Niket Patel, \texttt{nnp5656@nyu.edu}.}

\begin{document}

\maketitle



\section{Introduction}
\label{sec:intro}

Since the earliest work on the subject, mathematical reasoning has been a cornerstone of research in Artificial Intelligence (AI)~\citep{turing1939ordinal,newell1956logic}.
Recent work on applications of Large Language Models (LLMs) to solving mathematical conjectures has shown striking success, and we now have seen many instances of LLMs solving problems that have eluded human mathematicians~\citep{novikov2025alphaevolve,sothanaphan2026erdos728,alexeev2026shortproofs,alexeev2026shortproofsii,alon2026unitdistance,chen2026moonshine,oum2026cycledoublecover,huang2026sumproduct, openai2026navierstokes}.
Though LLM-based theorem provers can often make errors, they are particularly strong and reliable when their outputs are verified by a proof assistant such as Lean~4~\citep{moura2021lean4,mathlib2020,yang2023leandojo,xin2024deepseekprover,hubert2024alphaproof,ren2025deepseekproverv2}.
However, a common critique of applications of LLMs to solving mathematical conjectures is that they often only appear to solve problems that lie within the ``convex hull'' of existing mathematics.

Our long-term goal is to build systems capable of \emph{autonomous mathematical discovery}.
Such a system should be able to engage in \emph{open-ended} reasoning. Starting from a set of initial premises, the system should be able to formulate new questions, decide which are worth pursuing, construct and verify their proofs, and build on this knowledge~\citep{barkeshli2026structure}.
A fundamental challenge to achieving such a system is the question of interestingness of the mathematical statement.
The version of mathematics that was created by human intuition over thousands of years has proven to be an ``unreasonably effective'' tool to approaching real world problems~\citep{wigner1960unreasonable}. In contrast, arbitrarily adding trivial statements, or even nontrivial but \emph{uninteresting} statements, is unlikely to lead to the same success that mathematics has achieved in the past.

{
\begin{quote}
    { \emph{``Science is built up with facts, as a house is with stones. But a collection of facts is no more a science than a heap of stones is a house.''}} -- \emph{Science and Hypothesis,} \cite{poincare1905science}
\end{quote}
}

We can reason by analogy with Borges's ``Library of Babel''~\citep{borges1962library,litt2026babel}.
Borges's library holds every book that can be written, and therefore holds every truth in it; it is nonetheless useless, since no reader could find meaning within its nearly infinite shelves.
The space of all true mathematical statements has the same character, whereas mathematics, as it has been created by humans, does not.
Extrapolating outside the ``convex hull'' of mathematics, therefore, requires both the ability to navigate the space of possible deductions and a notion of value that distinguishes a discovery from a merely valid statement.

We seek a system that can begin with the mathematics that is already known, and recursively produce new mathematical results.
Any such autonomous system needs a quantitative object to optimize.
Prior work has often focused on ``Intrinsic Motivation'' \citep{poesia2024minimo, tsoukalas2025fermat}.
We take a different view, one in which there are inherent aspects of mathematics that can quantify the quality of a mathematical statement, its {\em interestingness}.
Prior work has found that LLMs do not robustly capture the same notions of interestingness, and the same diversity in those notions, as humans do~\citep{mishra2025matter}.
In the present work, we consider two complementary sources of value. On one hand, a theorem may be \emph{intrinsically interesting}, if it is easy to state but very difficult to prove.
Alternatively, a theorem could be \emph{useful} because of its relation to the surrounding body of mathematics.
A useful theorem is a reusable abstraction that -  once added to a body of mathematics -  compresses later proofs by making them easier to derive.

A fundamental primitive required to understand or estimate either of the above quantities is the difficulty to prove a theorem $T$, conditioned on a set of premises $P$ which are already available, which we will term $V(T|P)$.
Proof assistants \citep{moura2021lean4} allow a way to quantify the difficulty of a proof, in terms of the number of lines of code required to formally prove it. 
Lean~4's library \Mathlib provides a large dependency graph of machine-checked definitions, theorems, and proofs~\citep{mathlib2020}, which we can utilize to learn to model the conditional proof difficulty $V(T\mid P)$, mathematical interestingness and utility, and autonomous mathematical discovery. 
We make the following primary contributions.

\paragraph{Contributions.} 
\begin{enumerate}[leftmargin=1.4em]
  \item We formalize proof difficulty as the computational cost of deriving a target theorem from a given mathematical context in Lean 4. We post-train a 27B parameter LLM to predict the difficulty of a proof given a set of premises, and show that this predictor outperforms frontier LLMs.

  \item We motivate and define a notion of a theorem's \emph{interestingness} as the ratio between the length of the proof and the length of the statement of the theorem. We also define a notion of \emph{utility} based on the amount a theorem is able to compress a set of theorems and their proofs, and show that utility strongly correlates with interestingness.

  \item With our notion of \emph{interestingness}, we are able to show that we can post-train a language model to produce more interesting theorems, and we find that this quadruples the mean interestingness, as measured by the length of the real proof divided by the length of the statement. This training procedure also reduces the fraction of generated statements judged substantially or fully contained in \Mathlib from $91.9\%$ to $30.6\%$, showcasing the creation of more out-of-distribution math.
  
  \item We showcase our method in a specific setting where we apply our interestingness metric as an inference-time pruning algorithm to perform recursive mathematical discovery, and find nontrivial statements that are not present in \Mathlib.

\end{enumerate}

\section{Modeling the Conditional Proof Difficulty}
\label{sec:difficulty}

We ultimately want to train a model that can predict the difficulty of a proof without needing to write the proof beforehand. 
We model this difficulty as a function of both the target theorem and the premises available in the local context. 
We define the estimator of the difficulty of proving a theorem $T$ given premises $P$ as the value $V(T \mid P)$.
Such a premise-conditioned value function should satisfy certain properties.
For any codebase $\mathcal D$ of theorems and proofs stated in a formal language like Lean 4, we will denote by $C_\mathcal D(T|P)$ the ground truth number of lines of code in the codebase to prove theorem $T$ given premises $P$.
For any such codebase, this quantity is not necessarily defined for all choices of $T,P$, as some theorems may have multiple proofs, so we will define $V$ as satisfying the following conditions:

\begin{definition}[Conditional-value axioms]
\label{def:axioms}
A conditional proof computational-cost $V$ is \emph{faithful} on $\mathcal D$ if for all
targets $T \in \mathcal D$, lemmas $L \in \mathcal D$, and premise sets $P,Q \subset \mathcal D$, such that $P \subseteq Q$, the property (A1) below holds whenever $C_\mathcal D$ is well defined and (A2)--(A3) always hold:
\begin{align}
  \text{\textup{(A1) grounding:}}\quad
    &V(T\mid P)=  C_\mathcal D(T\mid P), \label{eq:grounding}\\
  \text{\textup{(A2) premise monotonicity:}}\quad
    &V(T\mid Q)\;\leq\;V(T\mid P), \label{eq:monotonicity}\\
  \text{\textup{(A3) composition:}}\quad
    &V(T\mid P)\;\leq\;V(L\mid P)+V(T\mid P\cup\{L\}).
    \label{eq:composition}
\end{align}
\end{definition}
Axioms (A2)--(A3) hold for the ideal shortest proof.
We would expect to see equality in (A3) if a lemma $L$ is required as an intermediate step towards proving the theorem from the premises, giving us the following Bellman-type relation:
\begin{equation}
  V(T\mid P) = V(L\mid P)+V(T\mid P\cup\{L\}).
  \label{eq:bellman}
\end{equation}
We will use these properties to inform the training objective in the following sections.

\subsection{Generating a Dataset}
\label{sec:data}
We use \Mathlib~\citep{moura2021lean4,mathlib2020} as a ground truth library from which we generate our training data.
For each theorem we extract its rendered type, the local definitions required to state it, the proof source, and the premises required for the proof.
If we were to naïvely train on the dependency graph of \Mathlib we would not get a useful predictor, as the library is built in a very atomized way.
But we still need to assign a label to each theorem $T$.
For instance, the median proof length in \Mathlib is $3$ lines of code, and the median number of times any particular premise is used is $1$.
To remedy this, we employ a technique we call \textbf{premise expansion}. 
We call a statement $L$ a premise of $T$ if it is \emph{directly} used in the proof of $T$, and we will write $P$ or $P(T)$ to denote the set of premises of $T$.
In this case, one \emph{expansion step} will remove $L$ from the set of premises of $T$, and add , $P(L)$, the premises of $L$ to the premises of $T$. So the new set of premises of $T$ is $P(T) \cup P(L) \backslash L$. We then add the number of lines of code in the proof of $L$ to the label of $T$.
Iterating moves the visible frontier backward through the dependency Directed Acyclic Graph (DAG)(Appendix~\ref{app:extraction}), creating a richer dataset of theorem, premises, proof triplets. We end up with around 100k data points which we use in the next section.
We evaluate on a withheld evaluation set coming from the same construction. 

\begin{figure}[t]
  \centering
  \includegraphics[width=\linewidth]{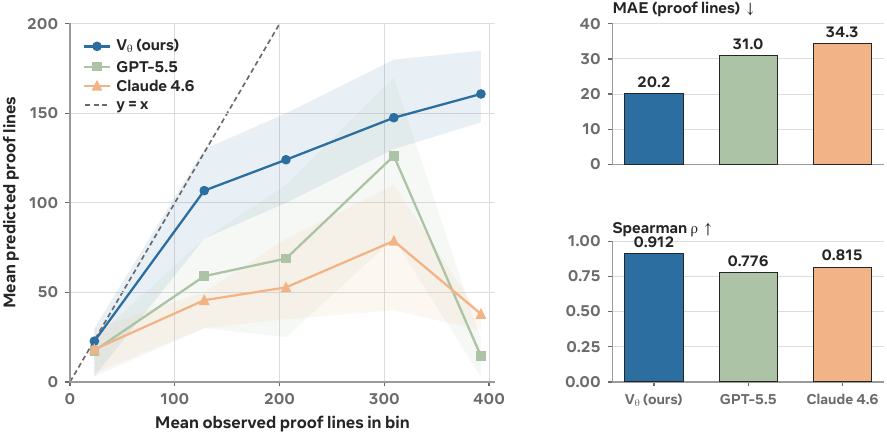}
  \caption{\textbf{The trained value is substantially better calibrated on held-out samples from \Mathlib.}
  The left panel groups examples by observed proof length, points give the mean observed and mean predicted proof lines, and shaded bands the interquartile range of predictions within each bin. The dashed line represents what we would expect from an optimal predictor. Note that all models underestimate proof length with increased length. The right plots report matched MAE and Spearman $\rho$ on all 4,615 validation prompts.}
  \label{fig:difficulty-results}
\end{figure}

\subsection{Modeling the Conditional Proof Length}
\label{sec:learning}

We want to train a model $V_\theta$ that can predict the difficulty of proving a theorem $T$ conditioned on a set of premises $P$. 
We fine-tune Qwen3.6-27B~\citep{qwen2026qwen36} with group relative policy optimization (GRPO)~\citep{shao2024deepseekmath, miles2026} on the dataset from Section~\ref{sec:data} for 350 steps.
Full definitions and specific details on the training setup are relegated to Appendix~\ref{app:optimization}.
We create a reward that is designed to satisfy the desiderata in Definition \ref{def:axioms}.
We design $\mathcal L_{\rm truth}$ to satisfy \eqref{eq:grounding}. If $L$ is a lemma used to go from $P$ to $T$, in order to satisfy \eqref{eq:bellman}, we introduce $\mathcal L_{\rm Bellman}$. To satisfy \eqref{eq:monotonicity}, we add $\mathcal L_{\rm drop}$, where $Q \subsetneq P$. Our final reward is,
\begin{align*}
    \mathcal R = &- |\log V_\theta(T|P) - \log C_\mathcal D(T|P)| 
        && (\mathcal L_{\rm truth}) \\
    &- |\log V_\theta(T|P) - \log( V_\theta(L|P) + V_\theta(T|P\cup\{L\}))| 
        && (\mathcal L_{\rm Bellman}) \\
    &- \max \{0, \log V(T|P) - \log V(T|P \backslash Q) \} 
        && (\mathcal L_{\rm drop})
\end{align*}

\label{sec:difficulty-results}
We evaluate models on a held-out validation set of 4,615 labeled validation prompts. All models receive identical definitions, premises, target, and output instruction.  We report mean absolute error (MAE) and Spearman rank correlation, computing both only over parsed non-negative answers (Appendix~\ref{app:evaluation}).
Figure~\ref{fig:difficulty-results} showcases performance of our trained model over GPT-5.5 ~\citep{openai2026gpt55systemcard} and Claude Opus 4.6 ~\citep{anthropic2026claudeopus46systemcard}. 
We find that while all three models tend to underestimate the difficulty of proofs, ours is substantially more accurate and better calibrated than the others.

\section{Interestingness-Driven Mathematical Discovery}
\label{sec:structure}

A human mathematician may find a mathematical conjecture valuable for many reasons.
They could be interested in a statement because of historical or social factors, like for instance that several famous mathematicians before them tried and failed to find a proof of a statement.
They could also study statements that are related to the physical world. These are factors which are extrinsic to the field of math, and rely on its relation to the world outside it. 
As such, these factors are not something that we could model in search of a quantitative definition of the interestingness of a statement; they lie outside the scope of our present objective.

We instead focus on structural properties that can be measured directly in a formal mathematical library.
In particular, statements that are simple to express but difficult to prove often align with human intuitions about mathematical interestingness.
Motivated by this observation, we define an intrinsic notion of interestingness that combines statement length with conditional proof difficulty. 
This quantity is not intended to capture every source of mathematical value, but to provide a concrete objective for guiding mathematical discovery.
We separately consider a theorem's utility, which measures its value to subsequent mathematical developments.

In Section \ref{sec:interestingness} we define and study the \emph{interestingness} of a mathematical statement.
In Section \ref{sec:utility} we examine the \emph{utility} of a mathematical statement and show its relation to the interestingness.
We then show in Section~\ref{sec:conjecturing} that interestingness is a concrete metric that can be optimized, via training and at inference-time.

\subsection{The Intrinsic Interestingness of a Mathematical Statement}
\label{sec:interestingness}

In this section, we attempt to define a notion of the interestingness of a mathematical statement or conjecture that aims to capture an \emph{intrinsic} property of the mathematical result, ignoring any relation to the outside world, or to other parts of the mathematical literature.
We claim that a statement $T$ is \emph{interesting} given a set of premises $P$ if it is easy to state, and has a long and incompressible proof.
To compute this, we can take a ratio between the length of a proof given some premises in terms of lines of code, and the number of characters of Lean 4 needed to state a theorem given some set of definitions one already has access to.
For instance, one would expect that a theorem in probability theory is easy to state for a reader who already has a measure-theoretic vocabulary and enormously difficult for a reader given access to only statements in algebraic geometry.
We recall the following quote by Pólya on the nature of aesthetics in math,

{
\begin{quote}
    { \emph{``The elegance of a mathematical theorem is directly proportional to the number of independent ideas one can see in the theorem and inversely proportional to the effort it takes to see them.''}} -- \emph{Mathematical Discovery, } \cite{polya1981mathematical}
\end{quote}
}

Formally, for any Lean declaration $X$, let $S(X)$ denote the number of characters in its statement, and let $C(X)$ be the set of all definitions required to state $X$, including whatever definitions are needed to state those definitions, recursively.
So $C(X)$ will contain all the definitions required to formally state $X$ from the axioms.
We will then define the conditional description length as
\begin{equation}
  L(T\mid P)\;=\;S(T)+\!\!\sum_{d\in C(T)\setminus C(P)}\!\!S(d).
  \label{eq:conditional-length}
\end{equation}

\begin{figure}[t]
  \centering
  \includegraphics[width=\linewidth]{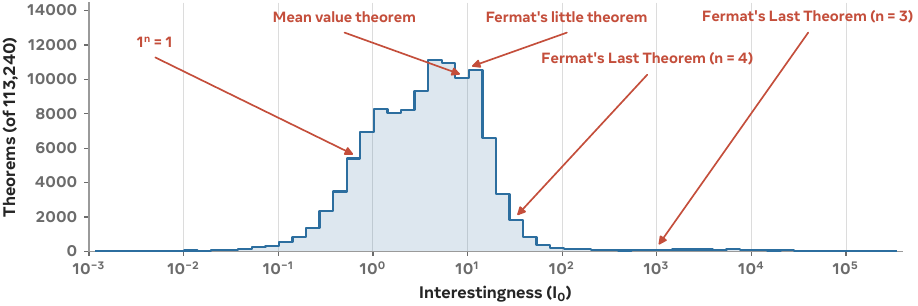}
  \caption{\textbf{Quantifying the interestingness of a statement.}
  Here we show the distribution of interestingness $I_0$, as defined in Section~\ref{sec:interestingness}, of all statements from \Mathlib.
  The resulting ordering is intuitive, with basic algebraic identities at the bottom, analysis in the middle because its definitional prerequisites are enormous, and theorems that are famously easy to state but hard to prove, like Fermat's Last Theorem for exponent 3, at the top.}
  \label{fig:interestingness-model-free}
\end{figure}

In this way, the term $L(T|P)$ counts the description length of all the definitions needed to state the theorem $T$, given  access to all the definitions used in the premises $P$.
We then formally define the conditional interestingness\footnote{We use a factor of 100 in the definition here so that the Interestingness has a mean \~1, as in Figure~\ref{fig:interestingness-model-free} and Figure ~\ref{fig:forward-interestingness}.} of a statement as 
\begin{equation}
  I(T\mid P)\;=\;100\;\frac{V(T\mid P)}{L(T\mid P)}.
  \label{eq:interestingness}
\end{equation}

We choose this definition to quantify the intrinsic difficulty of a mathematical statement as it is trivially possible to construct arbitrarily difficult statements, if we do not normalize by the description length of the statement. For instance, suppose we had a fixed set of premises $P$ and a set of theorems $\{T_1, T_2, \ldots\}$ that are all entirely unrelated and have a fixed proof length. Then the ``theorem'' $\hat T_n = T_1 \wedge T_2 \wedge \cdots \wedge T_n$ would have difficulty $V(\hat T_n | P) = O(n)$, whereas the interestingness is $I(\hat T_n | P) = O(1)$.

 In the special case where we consider the empty set of premises, we can create an absolute scale of the interestingness of a mathematical statement $I_0(T) =I(T|\varnothing)$\footnote{This corresponds to the notion of ``interest'' from \cite{aksenov2026compression}.}.
With no premises, the denominator is the target's own vocabulary length, $L(T\mid\varnothing)=S(T)+\sum_{d\in C(T)}S(d)$.
As we no longer require conditioning on a set of premises, the numerator can now be deterministically computed from a library such as \Mathlib, as we can just unroll the proof as written via premise expansion.
The result (Figure~\ref{fig:interestingness-model-free}) serves as a useful verification for the construction of our metric on existing statements of \Mathlib.
We can see in the bottom decile simple algebraic relations, like for instance the identity that $1^n = 1$, and in the upper quartiles we see results such as special cases of Fermat's Last Theorem, which happen to require very little machinery to state but are tough to prove.

\subsection{Intrinsic Interestingness Correlates with Extrinsic Utility}
\label{sec:utility}

\begin{figure}[t]
  \centering
  \includegraphics[width=\linewidth]{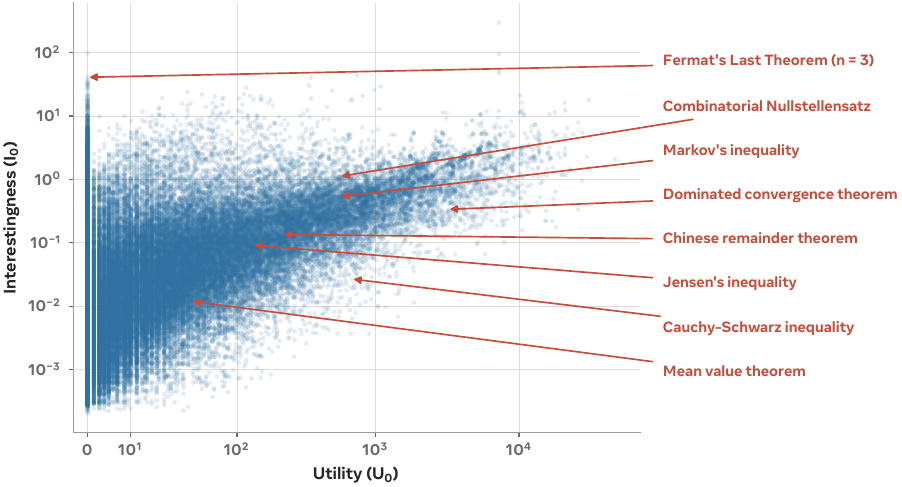}
  \caption{\textbf{Utility and interestingness are correlated.}
  Each point is a \Mathlib theorem. Utility $U_0$, defined in Section~\ref{sec:utility}, is the number of lines of code saved across the library when the theorem is admitted as a premise, while interestingness $I_0$, defined in Section~\ref{sec:interestingness}, is the ratio between a theorem's proof length and description length.
  Excluding the declarations with $U_0=0$  gives a Spearman $\rho=0.756$.
  }
  \label{fig:utility-vs-interestingness}
\end{figure}

We will define the \emph{utility} of a theorem to answer the following question: If we had access to this statement for free, how much would it compress the rest of the mathematical library?
Concretely, we compare two libraries: one in which $T$ is available as a premise, and one in which it is not, so that every statement relying on $T$ must re-derive it from scratch.
The utility of $T$ is the number of lines of code saved by the first library relative to the second.
If we let $D(T)$ be the set of theorems and lemmas that cite / use $T$ directly, then $\lvert D(T)\rvert$ would be the number of direct users of a theorem.
So we can define
\begin{equation}
  U_0(T)
    = \lvert D(T)\rvert\, \cdot V(T|\varnothing).
  \label{eq:utility-zero}
\end{equation}

Despite $U_0$ being an \emph{extrinsic} measure, depending on a theorem's relation to the broader mathematical theory surrounding it, and $I_0$ being an \emph{intrinsic} measure depending only on a theorem and its proof, we find that they are strongly correlated.
In Figure~\ref{fig:utility-vs-interestingness}, we see a strong positive correlation between $U_0$ and $I_0$ on \Mathlib, giving a Spearman correlation of $0.756$ if we disregard statements with no downstream users.
While some statements have high interestingness but low utility, few statements, if any, have high graph utility without high interestingness.
Interestingness is therefore a practical surrogate for the quality of a mathematical statement as it can be computed or estimated at proposal time, whereas utility is observable only after placing a theorem in a body of work and seeing its descendants. We recall the following quote of Thurston. 
{
\begin{quote}
    { \emph{``Our aesthetic instincts draw us to mathematics of a certain depth and connectivity. The very depth and beauty of the patterns makes them likely to be manifested, in unexpected ways, in other parts of mathematics, science, and the world.''}} --  \emph{Mathematical Education, }\cite{thurston2005mathematicaleducation}
\end{quote}
}

\subsection{Optimizing for Interestingness}
\label{sec:conjecturing}

\begin{figure}[t]
  \centering
  \includegraphics[width=\linewidth]{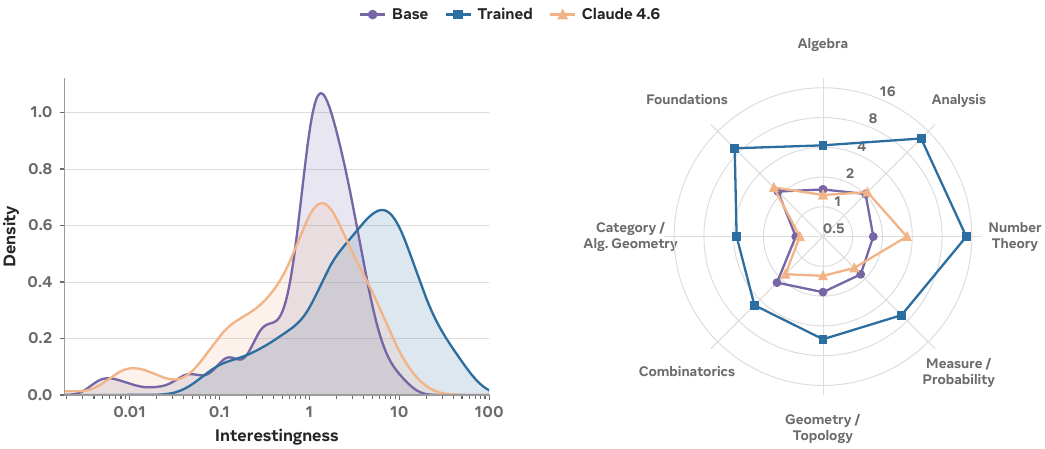}
  \caption{\textbf{Training shifts generated theorems toward higher ground-truth
  interestingness.}  Our trained 27B model is consistently able to produce more interesting statements, with 20 statements per coarse area.
  Left: kernel densities estimator (KDE; a method that smooths
the observed histogram with a kernel to give a continuous density estimate) fitted in $\log_{10}$-interestingness space over all 160 outcomes per model.
  Right: mean ground-truth interestingness by mathematical area over the same complete cohort, shown on a logarithmic radial scale from 0.5 to 20.  Colors identify the same models in both panels.}
  \label{fig:forward-interestingness}
\end{figure}

\begin{figure}[t]
  \centering
  \includegraphics[width=\linewidth]{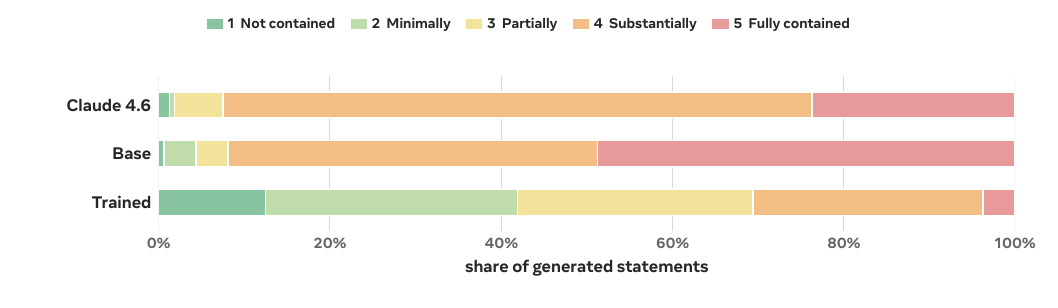}
  \caption{\textbf{Interestingness training reduces overlap with existing
  \Mathlib content.} For each model we take the 160 statements from Figure~\ref{fig:forward-interestingness}, and judge to what degree each statement is contained within \Mathlib. Appendix~\ref{app:containment} gives the rubric, prompt, and protocol.}
  \label{fig:mathlib-containment}
\end{figure}

The preceding construction turns the task of creating new, interesting, mathematics into an optimizable objective.
Given a mathematical context of prior lemmas and premises $P$, a conjecturing model, or \emph{conjecturer}, should be able to propose a statement $T$ which is cheap to express in the vocabulary of $P$ but requires a difficult proof.
With the use of our estimator $V_\theta$ we trained in Section~\ref{sec:difficulty}, we can obtain an estimate of the conditional interestingness, which we can use as a reward signal to train a model to produce more interesting conjectures.
From the \Mathlib training split we randomly select $10{,}000$ sets of premises. Each set $P$ contains at least 16 premises, with the median containing $77$. Appendix~\ref{app:forward-training} details the sampling procedure.
The model is conditioned on the available premises and definitions and emits a standalone Lean proposition. For a valid and nontrivial proposal, the training reward is,
\begin{equation}
  R(T,P)=0.25+\log\!\left(1+\frac{V_\theta(T\mid P)}{L(T\mid P)}\right)
  =0.25+\log\!\left(1+\frac{I_\theta(T\mid P)}{100}\right).
  \label{eq:conjecture-reward}
\end{equation}
Here $V_\theta$ is a frozen proof line count predictor and $L$ is the conditional description length from Equation~\ref{eq:conditional-length}.
The logarithm controls the influence of outliers while preserving the ordering induced by interestingness.
Parse failures receive reward $-0.5$, Lean-invalid or statements that don't relate to the premises at all receive $-0.25$, and statements which are able to be proven with simple proof automation tactics\footnote{For instance,  \texttt{assumption}, \texttt{rfl}, \texttt{simp}, \texttt{tauto}, and \texttt{simp\_all.}} receive zero. 
The constant term $0.25$ appears in the reward to provide a baseline reward for valid, relevant, and non-trivial statements.
Appendix~\ref{app:forward-training} describes elaboration, relevance, and triviality checks in full.

We evaluate Qwen3.6-27B~\citep{qwen2026qwen36}, trained with $75$ GRPO steps with the reward from \eqref{eq:conjecture-reward}, against the corresponding base model and Claude 4.6 on a held-out validation set from eight coarse mathematical areas in \Mathlib.
For each conjecturer and area, we repeatedly sample premises and generate candidate theorem statements, then ask a Claude Code with Opus 4.6 proving agent for an exact proof or a tightly constrained marginal repair.
Having a proof of each of these statements allows us to verify that they are correct, and to compute the ground-truth proof length, allowing us to get a ground-truth estimate of the interestingness. 
We retain $20$ proven statements per model and area, across 8 areas of mathematics, resulting in $160$ total statements per model. 
We showcase the ground-truth interestingness scores in Figure~\ref{fig:forward-interestingness};  full details are available in Appendix~\ref{app:forward-eval}.

We find that training produces a large shift in ground-truth verified interestingness (Figure~\ref{fig:forward-interestingness}).
Pooled across areas, mean $I(\cdot \mid P)$ rises from $1.76$ for the Qwen 3.6 27B base model, to $7.58$ for our trained model.
The trained model's mean is higher than the base model's in all eight areas, with area-level ratios ranging from $2.10\times$ in combinatorics to $8.72\times$ in number theory.
The trained model similarly beats Claude Opus 4.6 (prompted to generate interesting theorems) in all areas. 

The shift is also not explained merely by reproducing existing \Mathlib declarations (Figure~\ref{fig:mathlib-containment}).
With the conjecturer and subject area hidden, an LLM-as-a-Judge (in this case Claude Opus 4.6) rates each verified statement on the five-level containment rubric described in the caption.
Only $30.6\%$ of statements from the trained model are substantially or fully contained (scores 4--5), compared with $91.9\%$ for Base and $92.5\%$ for Claude.
Together, Figures~\ref{fig:forward-interestingness} and~\ref{fig:mathlib-containment} show that optimizing conditional interestingness moves the conjecturer toward statements with longer verified proofs relative to their descriptions and substantially less overlap with the existing library.

\subsection{Inference-Time Optimization for Theory Creation}
\label{sec:fwd}
\begin{figure}[t]
  \centering
  \includegraphics[width=\linewidth]{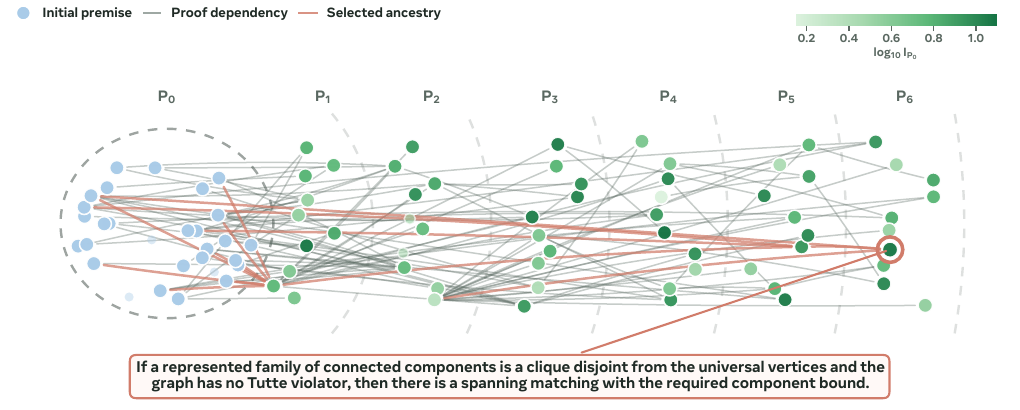}
  \caption{\textbf{Iterative forward discovery builds a reusable theorem graph.}
  Here we showcase a result from our iterative forward discovery algorithm introduced in Section \ref{sec:fwd}. 
  Green shade encodes $\log_{10}$ total interestingness relative to $P_0$ on a per-figure scale.
  The red lines trace the ancestry of the most interesting theorem introduced in $P_6$. More such graphs can be found in Figure \ref{fig:forward-dependency-graphs} in Appendix \ref{app:forward-eval}.}
  \label{fig:forward-dependency-graph}
\end{figure}

\begin{figure}[t]
  \centering
  \includegraphics[width=\linewidth]{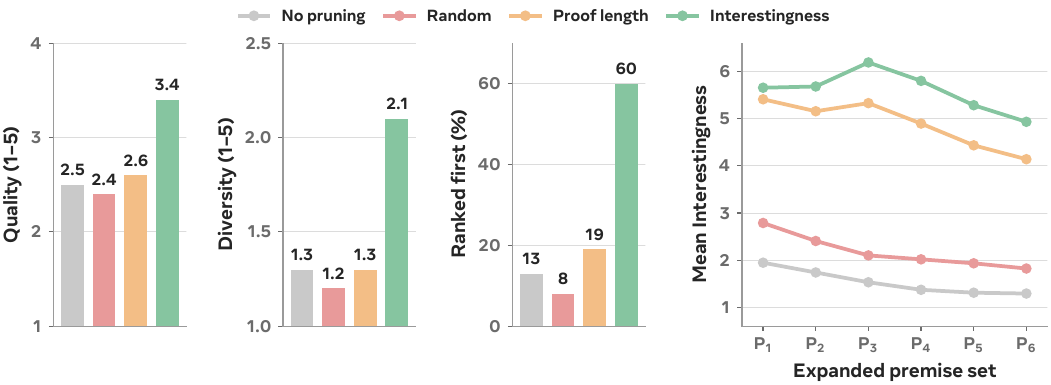}
  \caption{\textbf{Interestingness pruning is the most effective inference-time pruning criterion.} Left three panels: LLM judge scores over all 6 rounds for cohort quality and diversity (Table~\ref{tab:forward-pruning-judge}) and the share of four-way comparisons in which each rule's statement is ranked most interesting (Table~\ref{tab:forward-pruning-interest-judge}). Right: running mean of the verified interestingness of each set $P_n$ through $P_6$. More experiments are available in Appendix~\ref{app:inference-pruning}.}
  \label{fig:forward-pruning-summary}
\end{figure}

In the previous section, we showed that we can successfully train a conjecturer to generate more interesting individual conjectures. 
But our ultimate goal is not just to generate individual interesting theorems, but to iteratively build a body of interesting theorems, where one round's discoveries become the premises for the next round. 
We showcase here how we can use our metric of interestingness to achieve this at inference-time, without updating the weights of the model.
We evaluate an iterative theorem-discovery procedure, and we showcase that optimizing for interestingness gives us the strongest results across several metrics.
We want to have a system that starts from a set of premises $P_0$, and then adds verified theorems to create an iterated increasing set of premises $P_0 \subset P_1 \subset \cdots \subset P_N$.
At each iteration $n$, the conjecturer, here Claude 4.6, receives 20 premises, five of which are sampled from $P_{n-1} \backslash P_{n-2}$, and the rest from $P_{n-1}$. 
It generates 400 candidate statements. 
A Claude 4.6 based semantic filter removes equivalent conjectures and enforces diversity across theorem families. 
The surviving statements are then proved independently with Claude Code in Lean. 
Successfully verified statements are ranked by the interestingness as computed from the ground-truth proof, and the ten highest scoring results are ``promoted'' and added to the next set of premises $P_i$.
Appendix~\ref{app:inference-pruning} compares this rule against retaining every verified statement, promoting ten statements uniformly at random, and promoting the ten statements with the longest verified proofs; interestingness pruning yields the highest running mean of promoted-statement interestingness and the highest quality, diversity, and ``most interesting'' judgments (Figure~\ref{fig:forward-pruning-summary}).
We use the verified proof length here rather than $V_\theta$ because every candidate has already been proved at this stage, so the exact value of $V$ is available; the learned estimator is needed only where proving every candidate is too time-consuming, as in the reinforcement-learning setting of Section~\ref{sec:conjecturing}.
We showcase an experiment where $P_0$ is a set of premises from graph theory here in Figure~\ref{fig:forward-dependency-graph}; Appendix Figure~\ref{fig:forward-dependency-graphs} showcases the result of picking premises from Algebra, Measure / Probability Theory, and Number Theory.

The promotion rule at the end of each round is the only place where interestingness enters the loop, so we isolate its effect by holding the conjecturer, the semantic filter, and the proving agent fixed and varying only how the verified statements are pruned.
We compare four rules: retaining every verified statement, promoting ten uniformly at random, promoting the ten statements with the longest verified proofs, and promoting the ten with the highest ground-truth interestingness.
Each rule is run for 6 rounds from the same $P_0$, drawing on premises from graph theory, and we evaluate the resulting trajectories in two ways.
We aim to quantify the quality and diversity of the samples from each of the runs, which we are able to do by using an LLM-as-a-judge. 
The judge (Claude 4.6) is never told which promotion rule produced the statements it evaluates.
First, for each rule and round, it rates the cohort of ten promoted statements for quality and diversity on a 1--5 scale (Table~\ref{tab:forward-pruning-judge}).
Second, it is shown four statements at a time, one from each rule, and ranks them by how mathematically interesting they are (Table~\ref{tab:forward-pruning-interest-judge}).
We also track the running mean of the verified interestingness of the sets $P_n$, and we find that interestingness pruning is the strongest promotion rule on each of these metrics, as displayed in Figure~\ref{fig:forward-pruning-summary}.
Notably, pruning by raw proof length alone performs worse, indicating that the gain comes from the ratio defined in \eqref{eq:interestingness} rather than from simply favoring long proofs.

\section{Discussion}
\label{sec:discussion}

\label{sec:limitations}
Current LLM-based mathematics systems prove or solve human-posed problems; they do not decide which problems are worth posing.
Our work takes a step toward changing this by defining a simple notion of interestingness.
Crucially, our metric emerges from the structure of the proofs, and requires no human judgment of what constitutes interesting mathematics.
Our results show that this metric is already sufficient to steer a language model toward non-trivial, out-of-distribution conjectures.
Beyond individual conjectures, our iterative discovery procedure demonstrates that a system can build on its own outputs across rounds, with each round's most interesting theorems becoming premises for the next.
Together, these results offer a proof of concept that interesting self-expanding mathematical libraries, grown without human intervention beyond the initial premises, are feasible.

We identify the following limitations with our specific methodology in this paper.
Proof length in terms of lines of code can be a stylistic artifact, as automation tactics, such as \texttt{aesop} and \texttt{grind}, often trade proof length for runtime.
Alternative notions such as the ``Levin complexity'' might provide a richer signal in light of this \citep{livitanyi2008}.
Our definition also isolates some aspects of mathematical value, and one may object that much of what makes a theorem interesting is that it connects previously unrelated objects or unifies disjoint subfields.
Our definition of utility captures a weak form of this, but future work could try to quantify this notion more precisely.

Future work should aim towards extending our notions of discovery beyond just theorem statements to also allow for the creation of new and useful definitions, something we do not consider in this paper. 
Additionally, one could investigate training a model to estimate the proof difficulty in an online fashion, where as you expand your set of premises $P_0$, you are also updating your predictions of the proof length based on the ground truth outputs of your proving agent. 
Our results showed that LLMs are capable of creating a self-expanding mathematical library, which can have a high quality and diversity of theorems, but it is not yet clear if the new theorems are also useful for proving future results. 
Future work should look into the mechanisms for diversity and de-duplication that are required for long-running discovery systems.

\section*{Acknowledgments}
JK and NP thank the Simons Foundation for support through the Collaborative Grant “The Physics of Learning and Neural Computation” as well as support by the NSF through NRT Award 1922658. AH is supported by Hi! PARIS and ANR/France 2030 program (ANR-23-IACL-0005). NP thanks Vivien Cabannes, Charles Arnal, Taco Cohen, Anirudh Goyal, Skander Moalla and Anikait Singh for helpful discussions and feedback.

\bibliographystyle{assets/plainnat}
\bibliography{references}
\newpage
\beginappendix


\makeatletter
\let\apx@addcontentsline\addcontentsline
\renewcommand{\addcontentsline}[3]{%
  \apx@addcontentsline{#1}{#2}{#3}%
  \def\apx@type{#1}\def\apx@toc{toc}%
  \ifx\apx@type\apx@toc \apx@addcontentsline{atoc}{#2}{#3}\fi
}
\section*{\contentsname}
\@starttoc{atoc}
\makeatother
\newpage

\section{Related Work}
\label{sec:related}

\paragraph{Language models for formal theorem proving.}
Language models can generate formal proof steps~\citep{polu2020generative},
with subsequent systems adding tree search and reinforcement
learning~\citep{lample2022hypertree,gloeckle2024abel}, library retrieval
~\citep{yang2023leandojo}, synthetic data~\citep{xin2024deepseekprover}, subgoal
decomposition, and test-time search
~\citep{hubert2024alphaproof,ren2025deepseekproverv2,lin2025goedelprover,
wang2025kimina}. These methods primarily optimize completion of a fixed target
under a particular prover and budget, whereas we aim to explore open-ended reasoning. 

\paragraph{Autoformalization and formal libraries at scale.}
LLM autoformalization has progressed from statement-level translation
~\citep{wu2022autoformalization} and paired datasets
~\citep{azerbayev2023proofnet,ying2024lean} to large Lean developments from
mathematical textbooks~\citep{urban2026130klinesformaltopology,
wang2026m2fautomatedformalizationmathematical,gloeckle2026automatic}.
\textsc{AutoformBot} and \Atlas further demonstrate coordination and
verification across 26 books~\citep{rammal2026atlas}. Recent work has also
begun formalizing research-level results, such as the De Giorgi--Nash--Moser
theory for elliptic PDEs~\citep{armstrong2026degiorgi}. These results establish
the scalability of formal-library construction while exposing planning
problems from missing infrastructure and long-range dependencies.

\paragraph{Automated mathematical discovery and conjecturing.}
AM and Graffiti pioneered automatic concept and conjecture generation
~\citep{lenat1977am,fajtlowicz1988graffiti}, and interestingness itself became
an explicit research problem~\citep{colton2000interestingness}.
Graffiti's successors, notably TxGraffiti~\citep{davila2024txgraffiti}, have
produced conjectures that led to multiple published papers over a decade of
human--machine collaboration~\citep{davila2025txgraffiti10years}; however,
these systems operate on graph invariants and their heuristic filters are
intrinsically tied to that domain.
\citet{bengio2024aimathematician} propose an information-theoretic view
of mathematical interestingness, suggesting that valuable frameworks have small
description length while being close to many provable statements; our
definition can be seen as a computable, formal-library-grounded instantiation
of this principle.
More recent systems jointly learn conjecturing and proof in small axiomatic
domains~\citep{poesia2024minimo} or learn interestingness objectives for theory
formation~\citep{tsoukalas2025fermat}. Proposals for autonomous discovery add
novelty assessment, result selection, and closed-loop knowledge growth
~\citep{barkeshli2026structure}. Our objective supplies such a signal at formal-
library scale, grounding intrinsic interestingness in verified proof length and
relating it to a theorem's extrinsic utility. This separates the ability to
generate valid statements from the harder problem of selecting discoveries.

\paragraph{Description length and mathematical compression.}
Kolmogorov complexity measures shortest description length
~\citep{solomonoff1964,kolmogorov1965,livitanyi2008}, while proof length and
statement description length are distinct finite-corpus proxies. Resource-bounded
information~\citep{xu2020usable,blier2018description} and recent views of
mathematics as a proof hypergraph compressed by useful abstractions
~\citep{barkeshli2026structure,aksenov2026compression} motivate our
library-grounded quantities. Interestingness compares proof and description
length, while utility measures downstream proof compression; both are relative
to the available mathematical context. They are computable, finite-corpus
proxies rather than claims about uncomputable description complexity.

\section{Training the Difficulty Model \texorpdfstring{$V_\theta$}{V-theta}}
\label{app:difficulty-model}

\subsection{Data}
\label{app:extraction}

\subsubsection{Source corpus}

From our extraction of \Mathlib, we get 113,547 declarations. For each declaration we record the fully qualified name, rendered type, and file path; the file-local definitions required to interpret the target and premises; the project-local theorem declarations referenced by the accepted proof; the source proof span and its non-blank physical line count; and the dependency metadata used to change which premises are visible.

From each theorem we build a group and are able to compute our reward within such a group. Fix a root theorem $T$, let
$P$ be the premises currently visible to the model, and let $c$ be the
materialized computational-cost of that pair: the number of proof lines needed
to derive $T$ from $P$. Write $D$ for the file-local definitions shipped with
the prompt and $\Deps(\cdot)$, for the extracted dependencies of a declaration. Sample a cited lemma uniformly from the
theorem declarations among the visible premises,
\[
  L \sim \mathrm{Unif}\bigl(P)
\]
Let $\ell$ be the proof cost of $L$. Now expand it: replacing $L$ by its
own dependencies gives the expanded premise set $P^{-}$, and dropping each of
its premises independently gives the thinned premise set $P^{.8}$,
\[
  P^{-} = \bigl(P \setminus \{L\}\bigr) \cup \Deps(L),
  \qquad
  P^{.8} = \{\, p \in P^{-} : u_p < 0.8 \,\},
  \quad u_p \sim \mathrm{Unif}[0,1].
\]
A proof from $P^{-}$ can no longer cite $L$ and must derive it, so it costs
$c+\ell$. Writing $D_{\rm pre}$ for the definitions visible before the
expansion and $D(L)$ for those of $L$ alone, the edge emits six prompts
$(\text{target}\mid\text{definitions},\,\text{premises})$ together with their
observed computational-cost labels.

\begin{center}
\small
\begin{tabular}{@{}llp{0.42\linewidth}@{}}
\toprule
Role & Prompt $\mapsto$ Label & Meaning \\
\midrule
\role{base}             & $(T \mid D,\, P^{-}) \mapsto c+\ell$ & Predict the difficulty of proving $T$ when $L$ is not among the given premises; one would have to prove $L$ along the way, so the answer grows by its $\ell$ lines. \\
\role{remaining\_keep}  & $(T \mid D,\, P^{-}\cup\{L\}) \mapsto c$ & Predict the difficulty of proving $T$ when $L$ is handed back as a premise; it can be cited directly, so the answer returns to $c$. \\
\role{remaining\_remove}& $(T \mid D_{\rm pre},\, P) \mapsto c$ & Predict the difficulty of proving $T$ from the premises available before $L$ was expanded; the answer is again $c$, reached from a different set of premises. \\
\role{lemma\_context}   & $(L \mid D,\, P^{-}) \mapsto \ell$ & Predict the difficulty of proving $L$ itself from the premises the \role{base} row was given; this is exactly the $\ell$ lines that \role{base} had to add. \\
\role{lemma\_native}    & $(L \mid D(L),\, \Deps(L)) \mapsto \ell$ & Predict the difficulty of proving the same $L$ from the premises it was actually proved from in \Mathlib; the answer must again be $\ell$. \\
\role{base\_drop}       & $(T \mid D,\, P^{.8}) \mapsto \varnothing$ & Predict the difficulty of proving $T$ after each premise is dropped at random; it carries no label, since removing premises cannot make a proof shorter, and feeds only the one-sided penalty. \\
\bottomrule
\end{tabular}
\end{center}

All six rows remain adjacent during reward construction, so that the reward can be evaluated within a group without any cross-batch bookkeeping.  Roots, not individual prompt rows, determine the data
split, so variants of one theorem never straddle the train/validation boundary.

For each root, generation starts from $P=\Deps(T)$ and
$c=\Nlines(T)$ and runs three deterministically seeded random expansion walks.
After an edge is emitted, the walk continues from $(P^-,c+\ell)$. Each walk
reservoir-samples at most three eligible edges, stops after 300 expansion
attempts or once the visible premise set reaches 150 premises, and excludes prompts above
60,000 characters. The walks and reservoirs use seed 31; validation roots are
randomly selected with probability $p=0.01$.

The final dataset contains 18,318 training groups (109,908 prompts) and 923
validation groups (5,538 prompts).  Grounded labels have mean 92.3, median 49,
90th percentile 245, and maximum 500 proof lines.  

\subsubsection{Prompt}

Every model receives the same instruction template.  The frontier-model
evaluations permit free-form reasoning but require the final integer after
\texttt{\#\#\#\#}; training uses identical semantic content.

\begin{lstlisting}[basicstyle=\ttfamily\footnotesize,frame=single]
You are given Lean definitions, premises, and a target theorem statement.
Estimate the proof difficulty: the total number of proof lines required to
derive the target theorem from these premises.

The definitions are file-local definitions that the statement and premises
depend on. Use them only to understand what the statement means -- do NOT
judge difficulty based on them, and do not count them as proof lines.

Pay attention to exactly which premises are available. Do not infer
difficulty from the number of premises alone.

Do not draft the proof. Just estimate the difficulty from the dependency
structure. Give a concise estimate, then put the final answer on its own line:

#### <integer>

Definitions:
[numbered Lean declarations]

Premises:
[numbered Lean hypotheses]

Target theorem:
theorem target : [rendered type] := by

What is the total proof difficulty (number of proof lines)?
\end{lstlisting}

Hypothesis names are normalized positionally to \texttt{h1}, \texttt{h2}, and so
on; the target is renamed \texttt{target}; and any proof body after
\texttt{:= by} is stripped.  The model sees file-local definitions first, then
premises, then the target.

\subsection{Model and Optimization}
\label{app:optimization}

\subsubsection{Training objective}

Write $m(x)=\lgp{x}$, let $v_{rj}$ be sampled completion $j$ for role $r$,
and let $\bar v_r$ be the mean of the parsed completions for that role within
the batch; we abbreviate the six role means of \role{base},
\role{remaining\_keep}, \role{remaining\_remove}, \role{lemma\_context},
\role{lemma\_native}, and \role{base\_drop} as $\bar v_{\rm base}$,
$\bar v_{\rm keep}$, $\bar v_{\rm rem}$, $\bar v_{\rm ctx}$,
$\bar v_{\rm nat}$, and $\bar v_{\rm drop}$. Algorithm~\ref{alg:reward} gives
the per-completion reward used before GRPO normalization. Define the
Bellman residual $a(x;y,z)=|m(x)-m(y+z)|$ and the agreement residual
$q(x,y)=|m(x)-m(y)|$.

\begin{algorithm}[t]
\caption{\textsc{TrainingReward} --- exact per-completion reward.}
\label{alg:reward}
\begin{algorithmic}[1]
\Require completion $v=v_{rj}$ of role $r$; role means $\bar v_{\rm base}$,
  $\bar v_{\rm keep}$, $\bar v_{\rm rem}$, $\bar v_{\rm ctx}$,
  $\bar v_{\rm nat}$, $\bar v_{\rm drop}$; label $c_r$
\Ensure reward $\mathcal{R}_{rj}$
\If{$v$ does not contain exactly one \texttt{</think>} followed by exactly one
  terminal \texttt{\#\#\#\# non-negative-integer}}
  \State \Return $-2000$
\EndIf
\State $e_{\rm truth}\gets 0$ if $r=\role{base\_drop}$, else $q(v,c_r)$
\If{$r=\role{base}$}
  \State $e_{\rm add}\gets\tfrac14\bigl[
    a(v;\bar v_{\rm keep},\bar v_{\rm ctx})+a(v;\bar v_{\rm keep},\bar v_{\rm nat})$
  \Statex \hspace{5.9em}$
    +\,a(v;\bar v_{\rm rem},\bar v_{\rm ctx})+a(v;\bar v_{\rm rem},\bar v_{\rm nat})\bigr]$
  \State $e_{\rm drop}\gets[\,m(v)-m(\bar v_{\rm drop})\,]_+$
\ElsIf{$r=\role{remaining\_keep}$ or $r=\role{remaining\_remove}$}
  \State $e_{\rm add}\gets\tfrac12\bigl[
    a(\bar v_{\rm base};v,\bar v_{\rm ctx})+a(\bar v_{\rm base};v,\bar v_{\rm nat})
    \bigr];\quad e_{\rm drop}\gets0$
\ElsIf{$r=\role{lemma\_context}$}
  \State $e_{\rm add}\gets\tfrac13\bigl[
    a(\bar v_{\rm base};\bar v_{\rm keep},v)+a(\bar v_{\rm base};\bar v_{\rm rem},v)
    +q(v,\bar v_{\rm nat})\bigr];\quad e_{\rm drop}\gets0$
\ElsIf{$r=\role{lemma\_native}$}
  \State $e_{\rm add}\gets\tfrac13\bigl[
    a(\bar v_{\rm base};\bar v_{\rm keep},v)+a(\bar v_{\rm base};\bar v_{\rm rem},v)
    +q(v,\bar v_{\rm ctx})\bigr];\quad e_{\rm drop}\gets0$
\Else \Comment{$r=\role{base\_drop}$}
  \State $e_{\rm add}\gets0;\quad
    e_{\rm drop}\gets[\,m(\bar v_{\rm base})-m(v)\,]_+$
\EndIf
\State \Return $-0.50e_{\rm truth}-0.35e_{\rm add}-0.15e_{\rm drop}$
\end{algorithmic}
\end{algorithm}

\subsubsection{Hyperparameters}

\begin{center}
\small
\begin{tabular}{@{}ll@{\qquad}ll@{}}
\toprule
Parameter & Value & Parameter & Value \\
\midrule
Base model & Qwen3.6-27B & Global batch & 384 samples \\
Rollout batch & 48 prompts & Samples per prompt & 8 \\
Rollout temperature & 1.0 & Context limit & 131,072 \\
Prompt limit & 65,536 & Response limit & 8,192 \\
Optimizer & Adam & Learning rate & $10^{-6}$, constant \\
$\beta_1,\beta_2$ & $0.9,\,0.98$ & Weight decay & 0.1 \\
GRPO clip & $0.20\,/\,0.28$ & KL coefficient & 0.01 \\
Steps & 350 & Policy loss & per token \\
\bottomrule
\end{tabular}
\end{center}

\subsection{Evaluation Protocols}
\label{app:evaluation}

Evaluation is performed with greedy decoding and an 8,192-token response cap, to match training.
MAE, median absolute error, log-space $R^2$,
and Spearman $\rho$ are computed over all valid outputs. Every row from Claude and the trained model $V_\theta$ is successfully parsed, whereas GPT-5.5 produced 6 invalid outputs across all 4,615 prompts.

Calibration points in Figure~\ref{fig:difficulty-results} divide $[0,450]$ proof lines into five equal-width 90-line bins.
Within each observed-difficulty bin the marker gives the mean observed label against the mean prediction, and the shaded band spans the 25th--75th percentiles of predictions. 
Binning is used only for the plot on the left of Figure~\ref{fig:difficulty-results}; every aggregate metric on the right is computed on unbinned rows.

\section{Training a Conjecturing Model}
\label{app:conjecturing}

\subsection{Conjecturing Model Training}
\label{app:forward-training}

\subsubsection{Data construction}

The theorem-proposal data are reconstructed from the \role{base\_drop} rows of the corpus from \ref{app:extraction}, and only the premises are used here.
Premises are renamed positionally, and contexts with fewer than 16 premises or more than 12,288 tokens are removed. 
From this we randomly select 10,000 sets of premises for training and an additional 512 for validation, each set containing a median of 77 premises.
The model sees definitions and premise statements but never the source theorem or its proof. 
Section~\ref{app:proposal-prompt} gives the verbatim system and user templates.

\subsubsection{Compilation, relevance, and triviality}

A response must contain exactly one nonempty expression between
\texttt{<<<STATEMENT>>>} and \texttt{<<<END>>>}. Pantograph~\citep{aniva2025pantograph} compiles the expression as the type of a standalone theorem against the \Mathlib
environment, to ensure it is syntactically valid. We only allow for the model to produce theorems and reject declarations, definitions, proof commands,
\texttt{:=}, and \texttt{sorry}; parse failures receive $-0.5$ and expressions
that fail compilations receive $-0.25$.

For a compiled type, the global definition graph computes
$L(T\mid P)$ by subtracting the vocabulary exposed by the supplied premises.
The proposal must share at least one non-generic identifier with a
premise/definition context; failure receives $-0.25$. Direct premise restatements (as measured by token-level similarity) are marked trivial. Other candidates are tested for triviality using \texttt{assumption}, \texttt{rfl}, \texttt{simp}, \texttt{tauto},
and \texttt{simp\_all}; a candidate closed by any one receives reward of $0$. Only valid,
relevant, nontrivial statements are sent to the $V_\theta$ LLM to judge the difficulty, and they then receive reward according to Equation~\ref{eq:conjecture-reward}.

\subsubsection{Optimization configuration}

\begin{center}
\small
\begin{tabular}{@{}ll@{\qquad}ll@{}}
\toprule
Parameter & Value & Parameter & Value \\
\midrule
Initialization & Qwen3.6-27B & Steps & 75 \\
Samples per prompt & 8 & Rollout temperature & 1.0 \\
Global / rollout batch & 192 / 24 & Context / prompt limit & 12,800 / 12,288 \\
Response limit & 512 & Planned updates & 100 \\
Optimizer & Adam & Learning rate & $10^{-6}$, constant \\
$\beta_1,\beta_2$ & $0.9,0.98$ & Weight decay & 0.1 \\
GRPO clip & $0.20/0.28$ & KL penalty & $0$ \\
Save / evaluation interval & 5 / 25 & Policy loss & per token \\
\bottomrule
\end{tabular}
\end{center}

\subsubsection{Theorem-proposal prompt}

Bracketed fields such as \texttt{[NUMBERED PREMISES]} are populated mechanically
with the corresponding Lean declarations; they are not additional instructions.

\label{app:proposal-prompt}

The system message is:

\begin{lstlisting}[basicstyle=\ttfamily\scriptsize,frame=single,breaklines=true]
You are an expert Lean 4 and Mathlib theorem proposer. Answer immediately with the tagged theorem statement. Do not think aloud, explain, explore, or write a proof. Your first output must be <<<STATEMENT>>> and the whole response should stay under 256 tokens.
\end{lstlisting}

The user message is:

\begin{lstlisting}[basicstyle=\ttfamily\scriptsize,frame=single,breaklines=true]
Given the Lean premises below, propose one new, interesting theorem statement that is a plausible logical consequence of them. The statement will be scored by estimated proof difficulty divided by its premise-relative definition length.

A strong theorem is nontrivial, concise, and uses multiple premises in a meaningful way. It must be a well-typed Lean proposition that could actually be proved from the available premises. Do not output False, an unsupported conjecture, a restatement of one premise, a vacuous implication, or a reflexive identity such as `x = x` or `P <-> P`. Use mathematical objects or operations supplied by at least two different premises. The output must be a standalone proposition: do not mention local aliases such as `h1` in the generated statement.

Prefer a short consequence obtained by instantiating and chaining the exact premises over a broad theorem with many freshly invented binders. Copy namespaces, binder types, typeclass assumptions, and function argument order exactly from the declarations below. Do not guess shorthand names or notation that is not shown. If you use a premise name such as `h2`, apply it only at its displayed signature.

Silently type-check the final expression before emitting it: fully apply maps that need arguments, give every new variable an explicit type, and use each operation only on values of its declared type (for example, do not add or subtract propositions). The output is a type, so do not use `let`, `by`, `:=`, holes, tactics, proof terms, or `#` cardinality notation. Do not print this check.

# Definitions for understanding the premise vocabulary
[NUMBERED DEFINITIONS, WHEN PRESENT]

# Available premise declarations ([POSITIONAL PREMISE NAMES])
[NUMBERED PREMISES]

# Output contract
Output only a single standalone Lean type expression between the tags. Do not reference the premise aliases above. If new variables are necessary, bind them with a leading `forall`/`forall`. Do not include a theorem name, `:= by`, proof, code fence, or commentary.

<<<STATEMENT>>>
<one concise Lean theorem statement>
<<<END>>>
\end{lstlisting}

\subsection{Conjecturing Model Evaluation}
\label{app:forward-eval}

\subsubsection{Balanced verified cohort}

The eight strata we select premises from are Algebra, Analysis, Number Theory, Measure/Probability, Geometry/Topology, Combinatorics, Category/Algebraic Geometry, and Foundations.
For each group, we sample theorems from each of the three models we evaluate (Claude, the trained conjecturing model, and the base model), and pass it to a Claude Code with Claude Opus 4.6 as a proof agent. We repeat until we have 20 distinct theorems from each area.

The proving agent first attempts the statement byte-for-byte. If that fails, a separate
marginal-repair pass may add a necessary hypothesis, correct an index or bound,
or narrowly weaken a conclusion. A repair is accepted only if it compiles,
preserves the counts of $\forall,\exists,\leftrightarrow,\land,\lor,\neg$, passes
anti-vacuity checks, and has a token-set Jaccard similarity at least $0.70$, to disallow the statement becoming irrelevant or too easy.
We recompute proof lines and conditional length from the final compiled declaration and compiled proof, so the reported score is ground-truth interestingness under the proving pipeline rather
than the difficulty model's training reward.

\subsubsection{Mathlib containment}
\label{app:containment}

Claude Opus 4.6 at temperature 0
independently judges all 480 verified statements with proposer, area, and model
hidden. Each prompt includes the final statement, verified proof body, and
referenced \Mathlib declarations as non-exhaustive evidence. The proof scaffold
is pinned to \Mathlib revision
\texttt{905b95...}. Scores range from 1 (not contained) to 5 (an exact or trivially equivalent named
declaration exists); score 4 denotes a few-line wrapper or immediate
specialization. The judge must name a concrete declaration for score 5 and
returns structured JSON. Figure~\ref{fig:mathlib-containment} reports all of the recorded
statements per model.  The complete rubric and judge prompt are provided in
Section~\ref{app:containment-prompt}.

\subsubsection{Mathlib-containment rubric and judge prompt}

\label{app:containment-prompt}

The five-level rubric is:

\begin{lstlisting}[basicstyle=\ttfamily\scriptsize,frame=single,breaklines=true]
5 -- Fully contained. The exact statement (or a direct, trivially equivalent reformulation) exists in Mathlib as a named declaration. A user could cite it directly, or with a thin wrapper.

4 -- Substantially contained. The statement is provable in a few lines from existing Mathlib lemmas, or a strictly more general version is in Mathlib and the target is an immediate specialization.

3 -- Partially contained. Core ingredients (definitions and key supporting lemmas) exist in Mathlib, but the headline statement itself is not there and would require nontrivial assembly.

2 -- Minimally contained. Only background definitions or unrelated prerequisites are available; the substantive content is missing.

1 -- Not contained. Mathlib does not contain the statement or the necessary specialized definitions.

Operational guidance:
- Judge library containment, not mathematical validity, beauty, novelty, importance, or apparent difficulty.
- "A few lines" means a short Lean proof assembled directly from existing declarations, not merely a short informal mathematical argument.
- Use score 5 only when there is a concrete matching named declaration (possibly with symmetry, simplification, notation changes, or an immediate specialization). Identify that declaration whenever possible.
- A verified proof and its referenced Mathlib declarations are supplied as evidence. They are not exhaustive: a generated proof may overlook a more direct existing theorem, and proof length is not by itself dispositive.
- Do not infer a high score merely because the theorem is elementary. Conversely, do not infer a low score merely because its generated proof is long.
- If the exact declaration name is uncertain, report that uncertainty and avoid score 5 unless the API match is otherwise unmistakable.
\end{lstlisting}

Each item is judged with:

\begin{lstlisting}[basicstyle=\ttfamily\scriptsize,frame=single,breaklines=true]
You are an impartial expert judge of Lean 4 Mathlib library containment.

Apply the supplied rubric exactly. Estimate how much of the target theorem is already contained in Mathlib. The theorem has been verified, so do not judge whether it is true. Do not judge its mathematical novelty, importance, elegance, or difficulty.

The proof body and referenced declarations are supporting evidence, not an exhaustive Mathlib search. A generated proof can miss a direct existing theorem. Treat score 5 as requiring a concrete matching named declaration or an unmistakably direct API match. Distinguish a genuinely few-line Mathlib wrapper (score 4) from nontrivial assembly of ingredients (score 3).

<rubric>
[RUBRIC ABOVE]
</rubric>

<target_theorem>
[STATEMENT]
</target_theorem>

<verified_proof_body lines="[LINE COUNT]">
[PROOF BODY]
</verified_proof_body>

<referenced_mathlib_declarations>
[NAMES AND TYPES, OR NO RECOVERED LIST]
</referenced_mathlib_declarations>

Return exactly one JSON object and no Markdown or surrounding commentary:
{
  "score": <integer 1-5>,
  "matching_declaration": "<concrete Mathlib declaration for score 5, best candidate otherwise, or none>",
  "supporting_declarations": ["<zero or more important declaration names>"],
  "rationale": "<concise explanation tied to the rubric>",
  "confidence": "<high | medium | low>"
}
\end{lstlisting}

\newcommand{\genexheading}{\subsection}
\providecommand{\genexheading}{\section}
\genexheading{Selected Generations}
\label{app:generated-examples}

Each item gives the final compiled Lean statement followed by a mathematical
informalization. 

\begin{list}{}{%
  \setlength{\leftmargin}{0pt}%
  \setlength{\rightmargin}{0pt}%
  \setlength{\labelwidth}{0pt}%
  \setlength{\labelsep}{0pt}%
  \setlength{\itemindent}{0pt}%
  \setlength{\itemsep}{1.25em}%
  \setlength{\parsep}{0pt}%
  \setlength{\topsep}{0.75em}%
}

\item
\begin{samepage}
\begin{lstlisting}[basicstyle=\ttfamily\footnotesize,frame=single,breaklines=true]
∀ q : ℍ,
  q * q = -1 ↔ q.re = 0 ∧ ‖q‖ = 1
\end{lstlisting}
\noindent\textbf{Quaternionic square roots of minus one.}
A real quaternion squares to $-1$ if and only if its real component is zero
and its norm is one. Equivalently, the square roots of $-1$ are precisely the
unit purely imaginary quaternions.
\end{samepage}

\item
\begin{samepage}
\begin{lstlisting}[basicstyle=\ttfamily\footnotesize,frame=single,breaklines=true]
∀ z : UpperHalfPlane,
  ∃ n : ℕ, 0 < n ∧
    ‖Complex.exp
      (2 * ↑Real.pi * Complex.I * ↑z) ^ n‖ < 1 / 2
\end{lstlisting}
\noindent\textbf{Decay of the upper-half-plane exponential.}
For every $z$ in the complex upper half-plane, some positive power of
$e^{2\pi i z}$ has absolute value less than $1/2$. 
\end{samepage}

\item
\begin{samepage}
\begin{lstlisting}[basicstyle=\ttfamily\footnotesize,frame=single,breaklines=true]
∀ (n : Type _) [Fintype n] [DecidableEq n]
    (R : Type _) [CommRing R]
    (A : Matrix n n R),
  A * A = 1 ↔
    A.det * A.det = 1 ∧
      A.adjugate = A.det • A
\end{lstlisting}
\noindent\textbf{Involutory matrices and their adjugates.}
For a finite square matrix $A$ over a commutative ring, $A^2=I$ if and only if
$(\det A)^2=1$ and $\operatorname{adj}(A)=(\det A)A$.
\end{samepage}

\item
\begin{samepage}
\begin{lstlisting}[basicstyle=\ttfamily\footnotesize,frame=single,breaklines=true]
∀ κ : Cardinal,
  κ * κ = κ ↔
    κ = 0 ∨ κ = 1 ∨ Cardinal.aleph0 ≤ κ
\end{lstlisting}
\noindent\textbf{Multiplicatively idempotent cardinals.}
A cardinal is unchanged by squaring exactly when it is $0$, $1$, or infinite
(equivalently, at least $\aleph_0$).
\end{samepage}

\item
\begin{samepage}
\begin{lstlisting}[basicstyle=\ttfamily\footnotesize,frame=single,breaklines=true]
∀ z : UpperHalfPlane,
  ∃ w : UpperHalfPlane,
    w = -(1 : ℂ) / (z : ℂ) ∧
      ∀ n : ℕ, n > 0 ↔
        (↑n : ℂ) * (z : ℂ) ≠ 0
\end{lstlisting}
\noindent\textbf{Modular inversion in the upper half-plane.}
For every upper-half-plane point $z$, the modular inverse $w=-1/z$ is again in
the upper half-plane. Moreover, for every natural number $n$, positivity of
$n$ is equivalent to the nonvanishing of the complex product $nz$.
\end{samepage}

\item
\begin{samepage}
\begin{lstlisting}[basicstyle=\ttfamily\footnotesize,frame=single,breaklines=true]
∀ (R : Type _) [CommRing R],
  ∀ (W' : WeierstrassCurve.Jacobian R)
      (P : Fin 3 → R),
    W'.Equation (W'.neg P) ↔ W'.Equation P
\end{lstlisting}
\noindent\textbf{Negation preserves the Jacobian equation.}
For a Jacobian Weierstrass model over any commutative ring, negating a
projective point preserves the curve equation: $-P$ lies on the curve exactly
when $P$ does.
\end{samepage}

\item
\begin{samepage}
\begin{lstlisting}[basicstyle=\ttfamily\footnotesize,frame=single,breaklines=true]
∀ (K : Type _) [Field K] (f : Polynomial K),
  Function.Injective
    (fun g : Polynomial K => g * f) ↔
      f ≠ 0
\end{lstlisting}
\noindent\textbf{Injectivity of polynomial multiplication.}
Over a field, right multiplication by a polynomial $f$ is injective on the
polynomial ring if and only if $f$ is nonzero.
\end{samepage}

\item
\begin{samepage}
\begin{lstlisting}[basicstyle=\ttfamily\footnotesize,frame=single,breaklines=true]
∀ (K : Type _) [Field K] [NumberField K]
    (x : K),
  x ≠ 0 ↔
    ∀ w : NumberField.InfinitePlace K,
      (NumberField.InfinitePlace.mk
        (NumberField.InfinitePlace.embedding w)) x > 0
\end{lstlisting}
\noindent\textbf{Positivity at every infinite place.}
An element of a number field is nonzero exactly when its value at every
infinite place is strictly positive, where an infinite place evaluates through
the absolute value associated with its embedding.
\end{samepage}

\item
\begin{samepage}
\begin{lstlisting}[basicstyle=\ttfamily\footnotesize,frame=single,breaklines=true]
∀ (α : Type _) [GeneralizedBooleanAlgebra α]
    (a b : α),
  a = b ↔
    ∀ c : α, a \ c = b \ c ∧ c \ a = c \ b
\end{lstlisting}
\noindent\textbf{Equality via relative complements.}
Two elements $a,b$ of a generalized Boolean algebra are equal exactly when,
for every $c$, removing $c$ from them gives the same result and removing them
from $c$ also gives the same result.
\end{samepage}

\item
\begin{samepage}
\begin{lstlisting}[basicstyle=\ttfamily\footnotesize,frame=single,breaklines=true]
∀ (V : Type _) [Finite V] (G : SimpleGraph V),
  G.edgeSet = ∅ ↔
    ∀ s : Set V, G.IsClique s ↔ s.ncard < 2
\end{lstlisting}
\noindent\textbf{Empty graphs characterized by their cliques.}
A finite simple graph has no edges if and only if its cliques are precisely the
vertex sets containing fewer than two vertices.
\end{samepage}

\item
\begin{samepage}
\begin{lstlisting}[basicstyle=\ttfamily\footnotesize,frame=single,breaklines=true]
∀ (V : Type _) (G : SimpleGraph V) [Finite V],
  G.IsClique (Set.univ : Set V) ↔
    ∀ (M : Gᶜ.Subgraph),
      M.IsMatching ↔ M.verts = ∅
\end{lstlisting}
\noindent\textbf{Completeness through matchings in the complement.}
A finite simple graph is complete if and only if a subgraph of its complement
is a matching exactly when that subgraph has no vertices.
\end{samepage}

\item
\begin{samepage}
\begin{lstlisting}[basicstyle=\ttfamily\footnotesize,frame=single,breaklines=true]
∀ n : ℕ, n > 0 ↔
  ∀ f : Polynomial ℝ, f.natDegree ≤ n - 1 →
    ((∀ k < n,
      Polynomial.eval
        (Polynomial.Chebyshev.node n k) f = 0) ↔
      f = 0)
\end{lstlisting}
\noindent\textbf{Vanishing at Chebyshev nodes.}
A natural number $n$ is positive exactly when every real polynomial of degree
at most $n-1$ that vanishes at all $n$ Chebyshev nodes is the zero polynomial
(and the zero polynomial vanishes at those nodes).
\end{samepage}

\item
\begin{samepage}
\begin{lstlisting}[basicstyle=\ttfamily\footnotesize,frame=single,breaklines=true]
∀ (C : Type _) [CategoryTheory.Category C]
    [CategoryTheory.Abelian C]
    (X Y : C) (f : X ⟶ Y),
  CategoryTheory.Mono f →
    (CategoryTheory.IsIso f ↔
      ∀ (Z : C) (g : Y ⟶ Z),
        (g = 0 ↔ f ≫ g = 0))
\end{lstlisting}
\noindent\textbf{Detecting isomorphisms in an abelian category.}
Let $f:X\to Y$ be a monomorphism in an abelian category. Then $f$ is an
isomorphism exactly when precomposition with $f$ detects zero morphisms out of
$Y$: for every $g:Y\to Z$, one has $g=0$ if and only if $g\circ f=0$.
\end{samepage}

\end{list}

\section{Inference-Time Pruning Ablation}
\label{app:inference-pruning}

In this experiment, we aim to isolate the effect of the promotion rule in the recursive discovery loop of
Section~\ref{sec:fwd}. All four trajectories we study start from the set of 80 premises $P_0$ from graph theory, and share the same Claude Opus 4.6 conjecturer model, and Claude Code based proving procedure. 
Each initial generation pass proposes 400 statements. Every prompt contains 20 premises: five sampled from the immediately preceding set $P_n \backslash P_{n-1}$ and 15 from the initial pool and older promoted rounds $P_{n-1}$.

These statements are proved independently in Lean. We compare four post-verification promotion rules: \emph{no pruning} retains every verified novel statement; \emph{random} retains ten uniformly at random; \emph{proof length} retains the ten statements with the longest verified proofs; and \emph{interestingness} retains the ten statements with the largest ratio
\begin{equation}
  \widehat I_{\rm ver}(T\mid P)
    =100\,\frac{\text{lines in the accepted proof of }T}
                    {\text{characters in the normalized statement of }T}.
  \label{eq:verified-forward-interestingness}
\end{equation}

\subsection{Comparison across promotion policies}

All policies share the same initial premises and differ only in which verified statements are promoted. Table~\ref{tab:forward-pruning-primary} compares the statements promoted over
the ten rounds. Mean interestingness is computed within each round and then
averaged across rounds, giving every round equal weight.

\begin{table}[H]
  \centering
  \small
  \begin{tabular}{@{}lrrrr@{}}
    \toprule
    Promotion rule  & Mean $\widehat I_{\rm ver}$ & Median $\widehat I_{\rm ver}$ & Mean proof lines \\
    \midrule
    No pruning       & 1.313 & 1.114 & 3.80 \\
    Random           & 1.556 & 1.175 & 4.39 \\
    Proof length     & 3.259 & 2.480 & \textbf{11.48} \\
    Interestingness  & \textbf{3.899} & \textbf{3.380} & 7.41 \\
    \bottomrule
  \end{tabular}
  \caption{\textbf{Quality of the promoted statements}}
  \label{tab:forward-pruning-primary}
\end{table}

Unsurprisingly, interestingness pruning produces the highest mean and median interestingness,
while proof-length pruning produces the longest proofs.
A direct advantage on promoted interestingness is partly expected because one
policy selects on that quantity. We therefore separately evaluate every unique
candidate successfully proved in Lean \emph{before} the current round's
promotion rule is applied. Table~\ref{tab:forward-pruning-candidates} reports
their number, mean and median interestingness, and mean proof length across all
ten rounds, using the same aggregation as
Table~\ref{tab:forward-pruning-primary}.

\begin{table}[H]
  \centering
  \small
  \begin{tabular}{@{}lrrrr@{}}
    \toprule
    Promotion rule  & Mean $\widehat I_{\rm ver}$ & Median $\widehat I_{\rm ver}$ & Mean proof lines \\
    \midrule
    No pruning      & 1.313 & 1.114 & 3.80 \\
    Random          & 1.461 & 1.271 & 3.74 \\
    Proof length    & 1.694 & 1.167 & \textbf{5.46} \\
    Interestingness & \textbf{2.332} & \textbf{1.754} & 4.56 \\
    \bottomrule
  \end{tabular}
  \caption{\textbf{Verified candidates before selection.}}
  \label{tab:forward-pruning-candidates}
\end{table}

The judge sees four statements at a time, one from each policy, and ranks them by mathematical interestingness. Policy names, proofs, and scores are hidden. We evaluate 100 such groups, and show the results below.

\begin{table}[H]
  \centering
  \small
  \begin{tabular}{@{}lrrr@{}}
    \toprule
    Promotion rule & Ranked first & Ranked in top two & Mean rank $\downarrow$ \\
    \midrule
    No pruning      & 13\% & 34\% & 2.89 \\
    Random          & 8\%  & 29\% & 2.95 \\
    Proof length    & 19\% & 64\% & 2.36 \\
    Interestingness & \textbf{60\%} & \textbf{73\%} & \textbf{1.80} \\
    \bottomrule
  \end{tabular}
  \caption{\textbf{Direct policy-blinded judgment of mathematical
  interestingness.}}
  \label{tab:forward-pruning-interest-judge}
\end{table}

The judge also evaluates ten groups of ten statements promoted by each policy in each round as a group. Policy names and scores are hidden. The judge rates the group's overall quality and diversity and identifies repeated theorem families.
For no pruning, which retains more than ten statements, we select ten at random.

\begin{table}[H]
  \centering
  \small
  \begin{tabular}{@{}lrrrr@{}}
    \toprule
    Promotion rule & Quality (1--5) & Diversity (1--5) & Families per ten & Redundant \\
    \midrule
    No pruning      & $2.50\pm0.53$ & $1.30\pm0.48$ & $2.60\pm0.97$ & $50.0\%\pm26.7\%$ \\
    Random          & $2.40\pm0.52$ & $1.20\pm0.42$ & $2.90\pm0.99$ & $53.0\%\pm18.9\%$ \\
    Proof length    & $2.60\pm0.52$ & $1.30\pm0.48$ & $2.90\pm0.74$ & $50.0\%\pm18.3\%$ \\
    Interestingness & $\mathbf{3.40\pm0.52}$ & $\mathbf{2.10\pm0.32}$ & $\mathbf{3.50\pm0.71}$ & $\mathbf{28.0\%\pm7.9\%}$ \\
    \bottomrule
  \end{tabular}
  \caption{\textbf{Policy-blinded cohort judgment.} Entries are mean $\pm$
  sample standard deviation across the ten round-level cohorts for each
  promotion rule.}
  \label{tab:forward-pruning-judge}
\end{table}

\subsection{Judge prompts}

The complete judge instruction is deliberately short and uses a broad notion
of what a mathematician might appreciate:

\begin{lstlisting}[basicstyle=\ttfamily\footnotesize,breaklines=true]
You are a mathematician comparing four formally verified theorem statements.

Rank them from most to least mathematically interesting. Use a broad notion of interestingness: a mathematician might appreciate a statement because it is nontrivial, surprising, elegant, conceptually illuminating, potentially useful or reusable, connects ideas, or opens further questions. Judge the mathematical content, not notation, verbosity, statement length, or guessed proof length.

[A]
<THEOREM>

[B]
<THEOREM>

[C]
<THEOREM>

[D]
<THEOREM>

Return exactly one JSON object and no Markdown:
{"ranking":["A","B","C","D"],"rationale":"one or two sentences","confidence":"high|medium|low"}
\end{lstlisting}

The complete cohort-judge prompt is:

\begin{lstlisting}[basicstyle=\ttfamily\footnotesize,breaklines=true]
You are an impartial expert judge of a collection of ten verified Lean 4 theorem statements.

The statements are all correct. Judge the collection itself, without guessing its source or generation method.

Rate two dimensions from 1 to 5.

MATHEMATICAL QUALITY
5: Mostly nontrivial, coherent, potentially reusable results that express substantive mathematical content.
4: Generally meaningful results with some routine or overly specialized items.
3: Mixed quality; several useful statements but substantial routine conjunctions, wrappers, or narrow variants.
2: Mostly shallow, over-specialized, or mechanically assembled statements.
1: Almost entirely trivial, incoherent, vacuous, or unhelpful statements.

SEMANTIC DIVERSITY
5: Many genuinely distinct theorem families or mathematical ideas, with little redundancy.
4: Several distinct families and only limited repetition.
3: A meaningful mixture, but multiple statements repeat the same core pattern.
2: Dominated by one or two narrow families with superficial variations.
1: Nearly all statements are equivalent, nested variants, or routine repackagings.

Group statements by their substantive conclusion and proof idea, not merely surface syntax. A stronger/weaker variant or conjunction of the same core facts should normally stay in the same family.

<statements>
[1]
<THEOREM>
...
[10]
<THEOREM>
</statements>

Return exactly one JSON object and no Markdown:
{
  "quality_score": <integer 1-5>,
  "diversity_score": <integer 1-5>,
  "distinct_families": <integer 1-10>,
  "redundant_statement_indices": [<zero or more integers 1-10>],
  "family_assignments": [{"indices": [<integers>], "family": "<short label>"}],
  "rationale": "<concise explanation>",
  "confidence": "<high | medium | low>"
}
\end{lstlisting}
\clearpage

\section{Additional Figures}

\subsection{Cross-area interestingness matrix}
\label{app:interestingness}

The displayed matrix uses ten targets from each of 12 areas.  Starting from one
area-conditioned premise frontier per column, we apply three synchronous
expansion rounds: every current extracted theorem premise is replaced by all
original theorem premises used in its proof, unavailable declarations remain
terminals, and each resulting frontier is deduplicated.  Fresh $V_\theta$
predictions are obtained for all 1,440 target--condition pairs.
Each target's score is divided by its same-area score before cell medians are
formed.  Only 28 off-diagonal 95\% confidence intervals exclude one.
Across the 132 off-diagonal cell medians, the median is $0.520$ and 73.5\% are
below one. 
We see here that, as we might expect, a definition-heavy field such as measure theory becomes much less interesting when it is not given premises from analytic areas. 
We find also that some areas, such as algebra, become more interesting when supplied with premises from other fields.

\begin{figure}[H]
  \centering
  \includegraphics[width=0.8\linewidth]{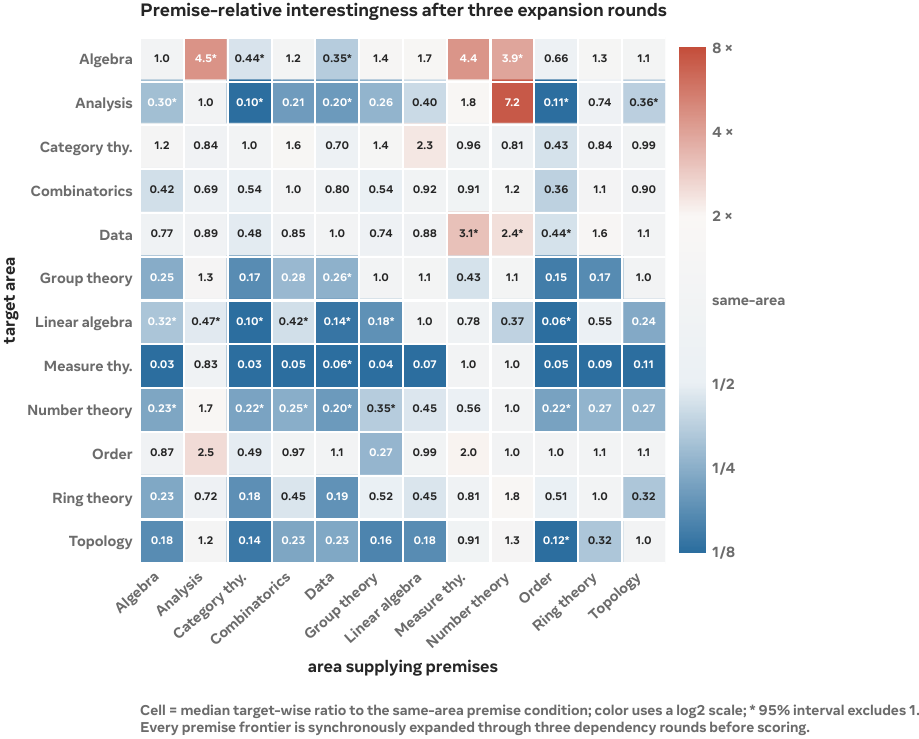}
  \caption{\textbf{Full cross-area interestingness matrix.} Each cell is the
  median target-wise ratio for the row area under premises from the column area
  versus the same-area premise condition. All premise frontiers are
  synchronously expanded through three rounds before new predictions are obtained. We select ten targets per area (120
  total); four row areas contribute nine ratios after each excludes one
  non-positive same-area score. Asterisks mark 95\% confidence intervals that exclude one.}
  \label{fig:area-matrix-full}
\end{figure}

\clearpage

\subsection{Iterative discovery graphs}

\begin{figure}[H]
  \centering
  \begin{subfigure}[b]{\linewidth}
    \centering
    \includegraphics[width=\linewidth]{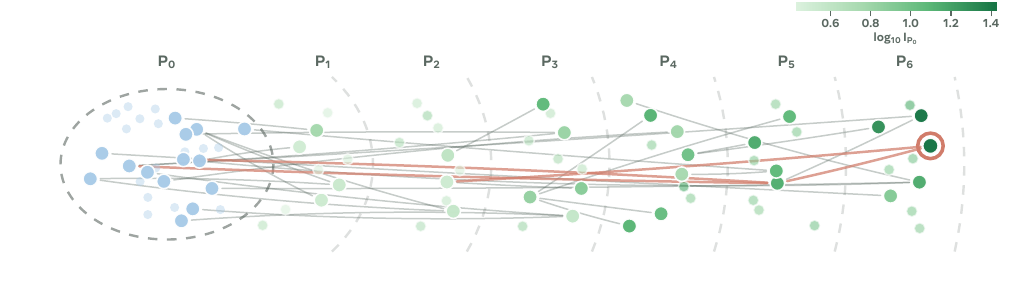}
    \caption{Algebra. For monic $q$ and $r$, reducing $p r$ modulo $q r$
    equals reducing $p$ modulo $q$ and then multiplying by $r$.}
  \end{subfigure}

  \medskip
  \begin{subfigure}[b]{\linewidth}
    \centering
    \includegraphics[width=\linewidth]{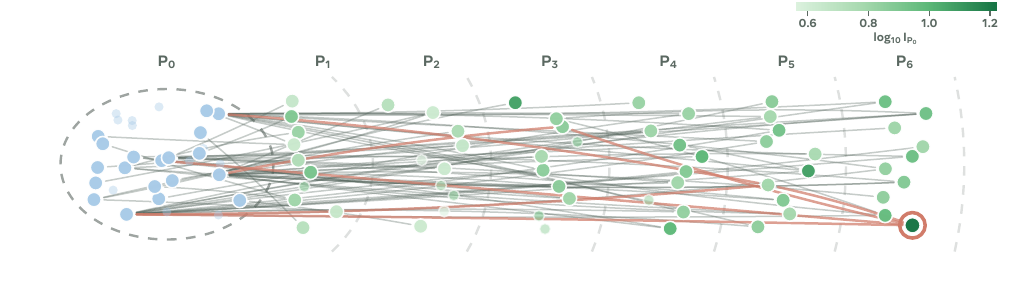}
    \caption{Measure/probability. If $\mu$ is absolutely continuous with
    respect to $\nu$, then under the stated measurability conditions, the sum
    of the pushforwards of $c\mu$ and $\mu$ is absolutely continuous with
    respect to $(c+1)$ times the pushforward of $\nu$.}
  \end{subfigure}

  \medskip
  \begin{subfigure}[b]{\linewidth}
    \centering
    \includegraphics[width=\linewidth]{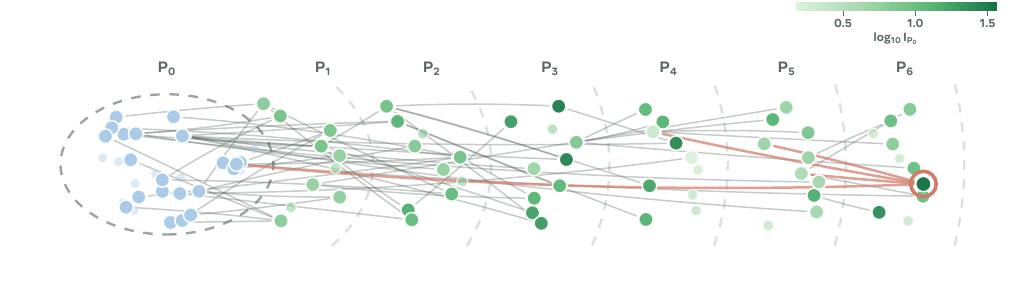}
    \caption{Number theory. Applying a real invertible $2\times2$ matrix to a
    nonreal complex number and then applying its inverse restores the imaginary
    part.}
  \end{subfigure}
  \caption{\textbf{Iterative forward-discovery graphs through $P_6$.}
  Each panel shows 30 of the 240 initial premises in $P_0$, always retaining
  every initial premise used by a displayed proof. Gray lines record explicit
  dependencies in accepted Lean proofs; coral marks the ancestry of the
  theorem with the highest total $P_0$-relative interestingness among the
  statements introduced in $P_6$. Green encodes $\log_{10} I_{P_0}$ with a
  separate scale in each panel. Each subcaption reports the selected statement.}
  \label{fig:forward-dependency-graphs}
\end{figure}

Figure \ref{fig:forward-dependency-graphs} provides more iterative discovery graphs.

\end{document}